\documentclass[12pt,journal,compsoc,onecolumn]{IEEEtran}
\ifCLASSOPTIONcompsoc
  \usepackage[nocompress]{cite}
\else
  \usepackage{cite}
\fi
\ifCLASSINFOpdf
\else
\fi
\RequirePackage{amsthm,amsmath,amsfonts,amssymb}
\RequirePackage[numbers]{natbib}
\RequirePackage[colorlinks,citecolor=blue,urlcolor=blue]{hyperref}
\RequirePackage{graphicx}
\theoremstyle{plain}

\newtheorem{theorem}{Theorem}[section]

\theoremstyle{remark}

\newtheorem{proposition}[theorem]{Proposition}
\newtheorem{assumption}[theorem]{Assumption}

\def\R{{\Bbb R}}

\def\E{{\Bbb E}}

\def\L{{\Bbb L}}

\def\H{\mathcal{H}}

\def\A{\mathcal{A}}

\begin{document}
%
\title{Kernel-based function learning in dynamic\\ and non stationary environments}
%
%
%
%

\author{Alberto Giaretta, Mauro~Bisiacco 
        and~Gianluigi~Pillonetto,~\IEEEmembership{Fellow,~IEEE}
\IEEEcompsocitemizethanks{\IEEEcompsocthanksitem A. Giaretta is with the Department of Pathology, University of Cambridge, Cambridge, UK. M. Bisiacco and G. Pillonetto are with the Department
of Information Engineering, University of Padova, Padova,
Italy.\protect\\
Corresponding author: Gianluigi Pillonetto (giapi@dei.unipd.it).}
}

%
%

\markboth{}%
{Shell \MakeLowercase{\textit{et al.}}: Learning functions in non stationary dynamic environments}
%



\IEEEtitleabstractindextext{%
\begin{abstract}
One central theme in
machine learning is function estimation 
from sparse and noisy data.
An example is supervised learning where
the elements of the training set are couples, each containing 
an input location and an output response.
In the last decades, a substantial amount of work has been devoted to
design estimators for the unknown function and 
to study their convergence
to the optimal predictor, also characterizing the learning rate. 
These results typically rely on stationary assumptions where input locations are drawn from 
a probability distribution that does not change in time.
In this work, we consider kernel-based ridge regression and derive convergence conditions
under non stationary distributions, addressing also cases where 
stochastic adaption may happen infinitely often. 
This includes the important exploration-exploitation
problems where e.g. a set of agents/robots has to monitor an environment to reconstruct
a sensorial field and their movements rules are
continuosuly updated on the basis of the acquired knowledge on the field and/or the surrounding environment. 
\end{abstract}

}

\maketitle

\IEEEdisplaynontitleabstractindextext

%
\IEEEpeerreviewmaketitle

\IEEEraisesectionheading{\section{Introduction}\label{sec:introduction}}


\IEEEPARstart{W}{e} consider the problem of reconstructing an unknown function $f$, defined over 
a set $X$, 
from noisy outputs $y_i$ collected over input locations $x_i$ \cite{Hastie01,Wahba:90,SpringerRegBook2022}. 
A typical measurements model is
\begin{equation}\label{MM}
y_i = \mu(x_i) + \nu_i, \quad i=1,\ldots,n 
\end{equation}
where the errors $\nu_i$ are zero-mean random variables
and $\mu$ is also called the regression function. Given any $x \in X$, 
it represents the optimal predictor
of future data in the mean squared sense.
In supervised learning, 
the input locations are also seen as random variables and 
the couples $\{x_i,y_i\}_{i=1}^n$ are realizations from a (typically unknown) probability 
distribution.  This particular version of the problem is also known as \emph{nonparametric 
regression under random design} in statistical literature \cite{Scholkopf01b}.\\
Many forms of regularization have been studied in the literature for supervised learning. 
Examples include spectral regularization methods, like early stopping and boosting \cite{Yao2007,Ref10,AravkinBoost,Blanchard2018},
and kernel-based regularization \cite{Scholkopf01b,PNAS2023} which will be the focus of this paper.
Within this framework, the function $f$ 
is assumed to belong to a Reproducing Kernel Hilbert Space (RKHS) $\mathcal{H}$,
a particular Hilbert space associated with positive-definite kernels \cite{Aronszajn50,Cucker01}.
Adopted estimators try to balance data fit and a regularizer built using  
the RKHS norm. They assume the form \cite{PoggioMIT}
\begin{equation}\label{Ja}
\hat{\mu}_n = \arg \min_{f \in \mathcal{H}}  \ \mathcal{J}(f(x_1),f(x_2),\ldots,f(x_n),\|f\|_{\mathcal{H}})
\end{equation}
with $J$ monotonically increasing w.r.t. the last argument
and $\|f\|_{\mathcal{H}}$ typically enforces smooth solutions.
Such formulation includes as special cases support vector machines 
\cite{Vapnik95,Cortes95,Drucker97} and
kernel ridge regression 
\cite{PoggioMIT,Scholkopf01b}. In particular, this last instance,
known in the literature also as regularization nework or regularized least squares \cite{Poggio90},
will be the focus of this paper. It arises when quadratic losses are used
to measure the discrepancy between the data $y_i$ and the predictions $f(x_i)$.\\
Kernel-based estimators \eqref{Ja} have been widely studied in the literature, with
many years of theoretical developments. 
One central theme is the derivation of conditions
under which $\hat{\mu}_n$ converges to the regression function $f$
as the data set size $n$ grows to infinity.
The literature on this subject is enormous and we cite e.g.  
\cite{Smale2007,Ref19,Girosi95,Ref27,LR2,Yao2007,Ref1,Ref11,Ref12,Blanchard2018}
where convergence rates and generalization properties of 
kernel-based estimates are obtained.\\  
A common feature of all of the above mentioned works is that 
the sampling distribution $p$ of the input locations is assumed time-invariant.
Hence, the input locations $x_i$ form a stationary stochastic process.  
An extension can be found in \cite{SmaleOnline} where $p$ is replaced 
with a convergent sequence of probability measures, so that stationarity of
$x_i$ holds asymptotically. In this paper we instead assume that 
the input locations distribution may vary freely in time over an arbitrary set $\mathcal{P}$. 
This point is important
in many engineering applications, e.g. in exploration-exploitation
problems and coverage control where a set of agents has to 
simultaneously explore the environment and
 reconstruct a sensorial field \cite{Schwager,Choi,Cortes:04,Bullo:09,Todescato2017}.
 Note that in this setting the stochastic mechanism underlying the input locations 
 establishes how the agents move inside the domain of interest.
 Hence, it is convenient to update 
 the movements rules 
on the basis of the acquired knowledge.
This latter can consist of the current filed estimate, 
which can suggest the most interesting areas to monitor,
or the agent could also follow external directives, independent of the field, 
requiring occasionally to inspect regions where an event is reported. 
The time-varying distributions might be Gaussian whose means and variances 
change over time on the basis of such directives.\\
Under this complex scenario, our aim is to derive conditions that ensure convergence 
of kernel-based estimators, in particular regularized least-squares, to the regression function. 
This will be obtained through a non trivial extension of the statistical learning 
estimates derived in \cite{Smale2007}. We will see that the convex hull  $co \mathcal{P}$ of the
set $\mathcal{P}$ plays a special role. 
In fact, the learning rate is related to 
smoothness of the unknown function and absolute 
summability of some covariances computed over $co \mathcal{P}$.\\

The paper is organized as follows. In Section \ref{Sec2}, 
we first recall the specific form
of the kernel-based estimator studied in the paper and state our assumptions
on the unknown function and the data generator.
In Section \ref{Sec3}, the main convergence result is reported
while Section \ref{Sec4} contains a numerical experiment.
Conclusions then end the paper.

\section{Preliminaries}\label{Sec2}

\subsection{Regularized least squares in RKHS}

Let $\H$ be the RKHS induced by the Mercer kernel $K: X\times X\rightarrow \R$,
with norm denoted by $\| \cdot \|_{\H}$. We consider kernel ridge regression given by 
the following specific form of the estimator \eqref{Ja}:
\begin{equation}\label{Tikhonov}
\hat{\mu}_t  = \arg \min_{f \in \H}  J_t(f) \\
\end{equation}
where
\begin{equation}\label{J}
J_t(f) = \frac{\sum_{i=1}^t \left( y_i- f(x_i) \right)^2}{t}     +   \gamma \| f \|_{\H}^2 
\end{equation}
where $\gamma$ is the so called regularization parameter which trades-off adherence to experimental data and
the RKHS norm which promotes smooth solutions. Differently from \eqref{Ja}, we now use $t$ in place of $n$ to denote the data set size to stress that a dynamic environment is now considered. In fact, the notation $t$ also indicates the (discrete-time) instant where the last measurement has been collected. In this setting, the input locations $x_i$ evolve in time and generally form a sequence of correlated random variables.

\subsection{Assumptions}
\label{Assum}

\indent {\bf{Kernel continuity}} The kernel $K$ inducing the RKHS $\mathcal{H}$ is assumed to be Mercer
(continuous) and the input space $X$ is a compact set
of the Euclidean space. Hence, 
there exists basis functions $\phi_j$ and positive scalars $\lambda_j$,
with $\sum_j \lambda_j<\infty$, providing 
the following absolutely convergent expansion
of $K$: 
\begin{equation}\label{Expansion}
K(x,a)= \sum_{j=1}^{\infty} \lambda_j \phi_j(x) \phi_j(a).
\end{equation}
Such decomposition is not unique and does
not necessarily arise from Mercer's theorem \cite{Minh:09}[Section 4.2]. 
We also assume that the kernel admits (at least) one particular expansion  where
all the $\phi_j$ in \eqref{Expansion} are
contained in a ball of the space of continuous functions. 
This condition is satisfied by the models commonly adopted in the literature like spline \cite{Bell:04},
Gaussian \cite{Minh:09}[Eq. 15]  and periodic kernels inducing sine and cosine eigenfunctions \cite{Ref27}. For interesting discussions around uniform boundedness 
of eigenfunctions coming from Mercer's theorem we also refer the reader to \cite{Zhou:02}.\\

\noindent {\bf{Regression function smoothness}} Regularity of $\mu$ is an important condition
entering convergence studies. 
Following the integral operator theory developed in 
\cite{Smale2007}, it can be stated making use of the important 
(integral) kernel operator. To introduce it, 
let $p$ be a probability density function on $X$
while $\L^2_p$ denotes the Lebesque space 
with the measure induced by $p$.
Hence, this is the space of real functions $f:X \rightarrow \R$ such that 
$$
\| f \|^2_p := \int_X f^2(a) p(a) da < \infty.
$$
Then, the kernel operator is 
$L_p : \L^2_p \rightarrow \H$,
and maps any $f \in \L^2_p$ into $h$ where 
\begin{equation}\label{BarL}
h(x):=L_p[f](x) = \int_{X} K(x,a) f(a) p(a) da, \quad x \in X.
\end{equation}
The level of smoothness of $h$ is now measured
computing the norm in $\L^2_p$ of $L_p^{-r}[h], \ r>0$.
The more we can increase $r$ maintaining finite 
the norm, the more regular the function $h$ is.
Our assumption on the regression function 
is then reported below. Beyond guaranteeing that $\mu \in \mathcal{H}$,
it also quantifies its level of regularity. 
The key difference w.r.t.
previous works is that our setting requires  that the smoothnes
condition be connected with all the probability densities 
$p$ in the convex hull  $co \mathcal{P}$.

\begin{assumption}[smoothness of the target function] \label{A2}
There exists $r$, with $\frac{1}{2}<r \leq 1$, and $\A_1$ with $0<\A_1  < \infty$, such that 
\begin{equation}\label{A2Cond}
\sup_{p \in co\mathcal{P}} \| L_p^{-r} \mu  \|_{p} < \A_1 .  
\end{equation}
\end{assumption} 
\begin{flushright}
$\blacksquare$
\end{flushright}
 
 {\bf{Data generation assumptions}} 
 We consider a very general framework to describe the process $\{x_i,y_i\}$
 in \eqref{MM} which evolves in time. 
Each input location  $x_i$ is  a random vector randomly drawn from 
a sequence of probability densities $p_i \in \mathcal{P}$.
We do not specify any particular stochastic or deterministic mechanism through which
these $p_i$ change over time. We just need an assumption on some covariances.
In fact, 
 beyond forming a non stationary 
 stochastic process, the input locations
 $x_i$ can be also correlated each other. Hence, a condition
 on the decay of the 
 covariance between a class of functions evaluated at different input locations is needed.
 In addition, we also need to assume that all the sampling distributions 
 that can be selected are Borel non degenerate w.r.t. the Lebesgue measure.
 This just implies that any selected $p$ allows the agent to visit all the input space $X$,
 a minimal condition to have consistency for any possible choice of $p$.
 This is summarized below.\\
 
\begin{assumption}[Data generation] \label{A1}
The probability densities $p \in \mathcal{P}$ are all Borel non degenerate w.r.t. the Lebesgue measure.
The errors  $\nu_i$ in \eqref{MM} are 
independent of each other and
of the $x_i$.
In addition, let $g$ be any function satisfying
$$
\| g\|_p < q < \infty, \quad \forall p \in co\mathcal{P}.
$$
Then, for every time instant $i$, there exists a constant  $\A_2 < \infty$, dependent on $q$ but 
independent of $g$, such that
\begin{equation*}
\sum_{k=0}^\infty   \left| \text{Cov}(g(x_i),g(x_{i+k})) \right| < \A_2  
\end{equation*}
where $\text{Cov}$ is the covariance operator.
\end{assumption}
\begin{flushright}
$\blacksquare$
\end{flushright}


\section{The main result}
\label{Sec3}

We are now in a position to report our main result. In its statement,  
the regularization parameter $\gamma$ becomes a function of $t$
which ensures statistical consistency of $\hat{\mu}_t $ in \eqref{Tikhonov}.
The learning rate is then discussed inside the proof.

\begin{proposition}\label{Main}
Let Assumptions \ref{A2} and \ref{A1} hold.  In addition, let the 
regularization parameter $\gamma$ depend on instant $t$ as follows
\begin{equation}\label{gamman}
\gamma \propto t^{-\alpha}, \quad 0<\alpha < \frac{1}{2}. 
\end{equation}
Then, as $t$ goes to infinity, one has
\begin{equation}\label{Result2}
\sup_{x \in X} |\hat{\mu}_t (x) - \mu(x) | \longrightarrow_p 0
\end{equation}
where $\longrightarrow_p$ denotes convergence in probability.
\end{proposition}

\subsection{Proof of the main result}

We show that, as $t$ goes to $\infty$, 
the estimator $\hat{\mu}_t$ 
converges in probability to $\mu$ in the topology of $\H$
and, hence, in that of the continuous functions \cite{Cucker01}.\\
Some notation is first introduced. For any time instant $t$, let $S_t: \mathcal{H} \rightarrow \mathbb{R}^t$ 
be the sampling operator defined by $S_t(f)=\{f(x_i)\}_{i=1}^t$ 
with its adjoint  $S^{\top}_t: \mathbb{R}^t \rightarrow \mathcal{H}$ given by
$S^{\top}_t c = \sum_{i=1}^t c_i K(x_i,\dot)$. It is well known, e.g. from \cite{Cucker01},
that this permits to formulate the minimizer of \eqref{J} as follows
\begin{equation}\label{UseofS}
\hat{\mu}_t  = \Big( \frac{S_t^{\top}S_t}{t} +\gamma I   \Big)^{-1}\frac{S_t^{\top}y^t}{t}
\end{equation}
where the vector $y^t$ contains the outputs up to instant $t$.
We also use
$$
p_1,\ldots,p_t 
$$
to indicate the first $t$ densities selected from $\mathcal{P}$. 
Repetitions could of course be present, like e.g. $p_1=p_2$.
The average density is
\begin{equation}\label{Barp}
\bar{p}_t(x) = \frac{ \sum_{i=1}^t p_i (x) }{t}.
\end{equation}
The nature of our problem requires now to introduce also
a sequence of Lebesgue spaces which vary over time.
In particular, $\L^2_t$ is the Lebesque space of real functions 
equipped with the measure induced by \eqref{Barp}.
It thus contains all the functions satisfying
$$
\| f \|^2_t := \int_X f^2(a)\bar{p}_t(a) da < \infty.
$$
Note that, in the description of the space and its norm, to simplify notation 
the integer $t$ in the
subscript replaces $\bar{p}_t$. The same convention is used also 
to introduce the sequence of kernel operators which  
map any $f \in \L^2_t$ into $h$ where
\begin{equation}\label{BarL}
h(x):= L_t[f](x) = \int_{X} K(x,a) f(a) \bar{p}_t(a) da, \quad x \in X.
\end{equation}
Instead, the operator $L_{p_i}$ is defined only by the sampling distribution at instant $i$, i.e.
it maps any $f \in \L^2_{p_i}$ into $h$ where
\begin{equation}\label{BarL}
h(x):= L_{p_i}[f](x) = \int_{X} K(x,a) f(a) p_i(a) da, \quad x \in X.
\end{equation}


The following function plays a key role in the subsequent analysis:
\begin{equation}\label{fgm}
\bar{\mu}_{t} = \arg \min_{f \in \H}  \| f-\mu \|^2_{t} + \gamma \| f \|^2_{\H}. 
\end{equation}
It is always well-defined since, by assumption, $\mu \in \mathcal{H}$ so that
it is continuous over $X$ and, in turn, belongs to $\L^2_t$ for any $t$.
Note that, differently from the data-free limit
function introduced in  
\cite[eq. 2.1]{Smale2007}, here $\bar{\mu}_{t}$
is a time-varying function, depending on the time instant $t$.
It could also represent a random function e.g. when the selection
of the sampling distributions $p_t$ are regulated by a stochastic mechanism.
In any case, exploiting the same arguments leading to \cite[eq. 2.2]{Smale2007}, 
the explicit solution of \eqref{fgm} is given by
\begin{equation}\label{fgmEx}
\bar{\mu}_{t} = (L_t+ \gamma I)^{-1}L_t \mu. 
\end{equation}
Consider now the following decomposition of the estimation error
\begin{equation}\label{DecomErr}
\|\hat{\mu}_t -\mu \|_{\H} \leq  \left \| \bar{\mu}_{t} -\mu \right \|_{\H} + \left \|\hat{\mu}_t -\bar{\mu}_{t} \right \|_{\H}
\end{equation}
We start analyzing the first term on the RHS of (\ref{DecomErr}).
The average density $\bar{p}_t$ varies over time but never escapes from 
$co \mathcal{P}$.  Then, combining
Assumption \ref{A2} and eq. (3.11) in \cite{Smale2007},
one obtains the following bound uniform in $t$:
\begin{equation}\label{FirstTerm}
\left \| \bar{\mu}_{t} -\mu \right \|_{\H} 
\leq    \gamma^{r-\frac{1}{2}} \| L_t^{-r} \mu  \|_{t} \leq \gamma^{r-\frac{1}{2}} \A_1. 
\end{equation}
To account for the possible stochastic nature of $\bar{\mu}_{t}$,
we can also take the expectation to obtain
\begin{equation}\label{FirstTermB}
\E \left \| \bar{\mu}_{t} -\mu \right \|_{\H} 
\leq    \gamma^{r-\frac{1}{2}} \| L_t^{-r} \mu  \|_{t} \leq \gamma^{r-\frac{1}{2}} \A_1. 
\end{equation}
Now, we study 
$\E \left \|\hat{\mu}_t -\bar{\mu}_{t} \right \|_{\H}$ which corresponds to the mean  
of the second term on the RHS of (\ref{DecomErr}).
Combining (\ref{Barp},\ref{BarL}) and \eqref{fgmEx}, 
one has 
\begin{equation*}
\gamma \bar{\mu}_{t} = L_t[\mu - \bar{\mu}_{t}] =\frac{1}{t}  \sum_{i=1}^{t} L_{p_i} [\mu - \bar{\mu}_{t}],    
\end{equation*}
while from \eqref{UseofS} one has
$$
\hat{\mu}_t - \bar{\mu}_{t} = \Big( \frac{S_t^{\top}S_t}{t} +\gamma I   \Big)^{-1}\Big(\frac{S_t^{\top}y^t-S_t^{\top}S_t \bar{\mu}_{t}-t \gamma \bar{\mu}_{t}}{t}\Big).
$$
Combining the last two equations, and taking into account the definitions of $S_t$ and of its adjoint $S_t^{\top}$,
we obtain
\begin{subequations}
\begin{align*}  \small
 & \left \|\hat{\mu}_t -\bar{\mu}_{t} \right \|_{\H}\\ \nonumber 
 & \leq \frac{1}{\gamma} \small  \left[  \left\| \frac{1}{t}  \sum_{i=1}^{t} \left((y_i- \bar{\mu}_{t}(x_i) ) K(x_i, \cdot)   
- L_{p_i}[ \mu - \bar{\mu}_{t}](\cdot)  \right )   \right\|_{\H} \right] 
\end{align*}
\end{subequations}
and this implies
\begin{subequations}
\begin{align*}  \small
 & \E \left \|\hat{\mu}_t -\bar{\mu}_{t} \right \|_{\H}\\ \nonumber 
 & \leq \frac{1}{\gamma} \small \E  \left[  \left\| \frac{1}{t}  \sum_{i=1}^{t} \left((y_i- \bar{\mu}_{t}(x_i) ) K(x_i, \cdot)   
- L_{p_i}[\mu - \bar{\mu}_{t}](\cdot)  \right )   \right\|_{\H} \right]. 
\end{align*}
\end{subequations}
To obtain the desired bound, we need to consider
\begin{equation}\label{SqNorm}
\E  \left[  \left\| \frac{1}{t}  \sum_{i=1}^{t} \left((y_i- \bar{\mu}_{t}(x_i) ) K(x_i, \cdot)  
- L_{p_i}[ \mu - \bar{\mu}_{t}](\cdot)  \right )   \right\|_{\H}^2 \right] 
\end{equation}
where the expectation takes into account all the randomness of the data $\{x_i,y_i\}_{i=1}^t$ and possibly also that underlying the operators $L_{p_i}$ since they depend on the choice of $p_1,\ldots,p_t$.
Exploiting the expansion \eqref{Expansion}, the kernel section centred on $x_i$ 
can be written as
$$
K(x_i,\cdot)= \sum_{j=1}^{\infty} \lambda_j \phi_j(x_i) \phi_j(\cdot)
$$
so that one has
\begin{eqnarray*}
(y_i- \bar{\mu}_{t}(x_i) ) K(x_i, \cdot) &=& \sum_{j=1}^{\infty} (y_i- \bar{\mu}_{t}(x_i) ) \lambda_j \phi_j(x_i) \phi_j(\cdot) \\
&=&   \sum_{j=1}^{\infty} a_j(x_i)   \lambda_j  \phi_j(\cdot) 
\end{eqnarray*}
where we have used the following correspondence
\begin{eqnarray*}
a_j(x_i) &=& (y_i - \bar{\mu}_{t}(x_i) )\phi_j(x_i)\\
&=& (\mu(x_i)+ \nu_i - \bar{\mu}_{t}(x_i) )\phi_j(x_i).
\end{eqnarray*}
Since the noises $\nu_i$ have all zero-mean and independent of the input locations $x_i$,
and recalling the definition of $L_{p_i}$,
taking the expectation we also obtain
\begin{eqnarray*}
\E\left[(y_i- \bar{\mu}_{t}(x_i) ) K(x_i, \cdot)\right] &=& \E\left[(\mu(x_i)- \bar{\mu}_{t}(x_i) ) K(x_i, \cdot)\right]\\
&=& L_{p_i}[ \mu - \bar{\mu}_{t}](\cdot)\\
&=&\sum_{j=1}^\infty \E[a_j(x_i)] \lambda_j \phi_j (\cdot).
\end{eqnarray*}

So, the terms inside the sum reported in \eqref{SqNorm} 
consist of differences between random functions and their means. 
Exploiting this result and also the RKHS norm's structure, 
we can now rewrite \eqref{SqNorm} as follows 
\begin{subequations}{\small
\begin{align*}  
&\frac{1}{t^2} \E  \left\|	 \sum_{i=1}^t \sum_{j=1}^\infty (a_j(x_i)-\E[a_j(x_i)])\lambda_j \phi_j(\cdot)	\right\|_{\H}^2 = \\
&= \E \left[ \sum_{i=1}^t \sum_{k=1}^t \sum_{j=1}^\infty \frac{\lambda_j}{t^2} \left(a_j(x_i) -\E[a_j(x_i)]\right)\left(a_j(x_k)-\E[a_j(x_k)]\right) \right]\\
&= \E \left[ \sum_{i=1}^t \sum_{k=1}^t \sum_{j=1}^\infty \frac{\lambda_j}{t^2} \left(\tilde{a}_j(x_i) -\E[\tilde{a}_j(x_i)]\right)\left(\tilde{a}_j(x_k)-\E[\tilde{a}_j(x_k)]\right) \right]
\end{align*}}
\end{subequations}
where
\begin{eqnarray*}
\tilde{a}_j(x_i) = (\mu(x_i) - \bar{\mu}_{t}(x_i) )\phi_j(x_i)
\end{eqnarray*}
and we have still used the fact that the noises are zero-mean and independent of the 
input locations. 
We now obtain an upper bound on the first term present in the rhs of the above equation.
First, evaluating the objective \eqref{fgm} at $f=\bar{\mu}_{t}$ and $f=0$, one has  
$$
\| \bar{\mu}_{t} - \mu \|_t \leq   \sqrt{\| \bar{\mu}_{t} - \mu \|^2_t + \gamma\| \bar{\mu}_{t} \|^2_{\mathcal{H}}}
 $$
 $$ 
 \leq \| \mu \|_t \leq  \sup_{p \in co\mathcal{P}}   \| \mu \|_p <\infty.
$$
where the last inequality derives from continuity of the function $\mu$ 
on the compact $X$. 
This, combined with the fact that (by assumption) all the $\phi_j$ are contained in a ball of the space of continuous functions, leads to the following bound, uniform in $t$ and $j$: 
$$
\| a_j(\cdot) \|_t \leq M < \infty. 
$$
Such inequality permits to exploit Assumption \ref{A1} to obtain
\begin{subequations}
\begin{align*}  \small
& \frac{1}{t^2} \E \left[ 	\sum_{i=1}^t \sum_{k=1}^t \sum_{j=1}^\infty  \lambda_j \left(a_j(x_i) -\E[a_j(x_i)]\right)\left(a_j(x_k)-\E[a_j(x_k)]\right) \right] \\
& \leq  \sum_{j=1}^\infty \frac{\lambda_j}{t^2} \sum_{i=1}^t \sum_{k=1}^t  \left | \E \left[ \left(a_j(x_i) -\E[a_j(x_i)]\right)\left(a_j(x_k)-\E[a_j(x_k)]\right) \right] \right| \\
 & \leq \frac{2 \A_2  \sum_{j=1}^{\infty} \lambda_j}{t}.  
\end{align*}
\end{subequations}
This last result, together with the Jensen's inequality, leads to 
\begin{equation}\label{Etai7}
 \E[\|\hat{\mu}_t -\bar{\mu}_{t} \|_{\H}  ]  \leq   \frac{\sqrt{2 \A_2  \sum_{j=1}^{\infty} \lambda_j }}{\gamma\sqrt{t} }. 
 \end{equation}
Combining (\ref{Etai7})  with  (\ref{DecomErr}) and (\ref{FirstTerm}), we obtain
\begin{eqnarray*} 
 \E[\|\hat{\mu}_t-\mu \|_{\H}] \leq \gamma^{r-\frac{1}{2}} \A_1  +  \frac{\sqrt{2 \A_2  \sum_{j=1}^{\infty} \lambda_j } }{\gamma\sqrt{t} }.
\end{eqnarray*} 
Using the suggested rule for updating the regularization parameter, 
i.e. $$\gamma \propto t^{-\alpha}, \quad 0<\alpha < \frac{1}{2},$$
 the convergence rate is
$$
\min\Big(\alpha (r-\frac{1}{2}),\frac{1}{2}-\alpha\Big)
$$
and this completes the proof.

\section{Numerical Experiment}\label{Sec4}

\begin{figure*}
\center {\includegraphics[scale=0.4]{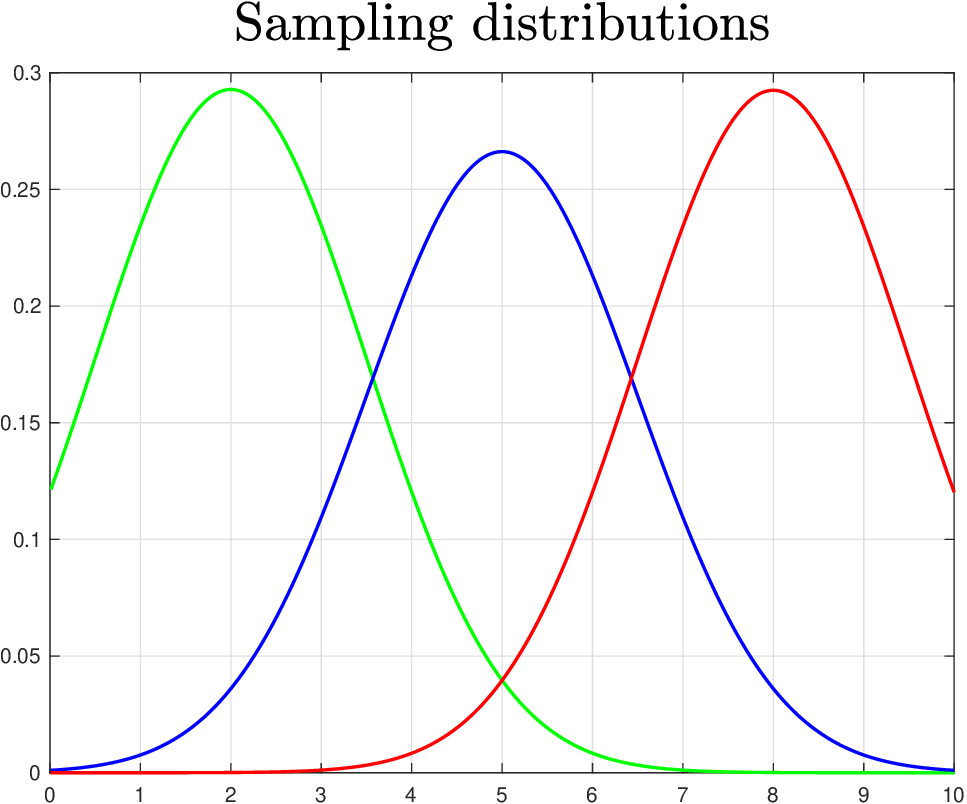}} \  {\includegraphics[scale=0.4]{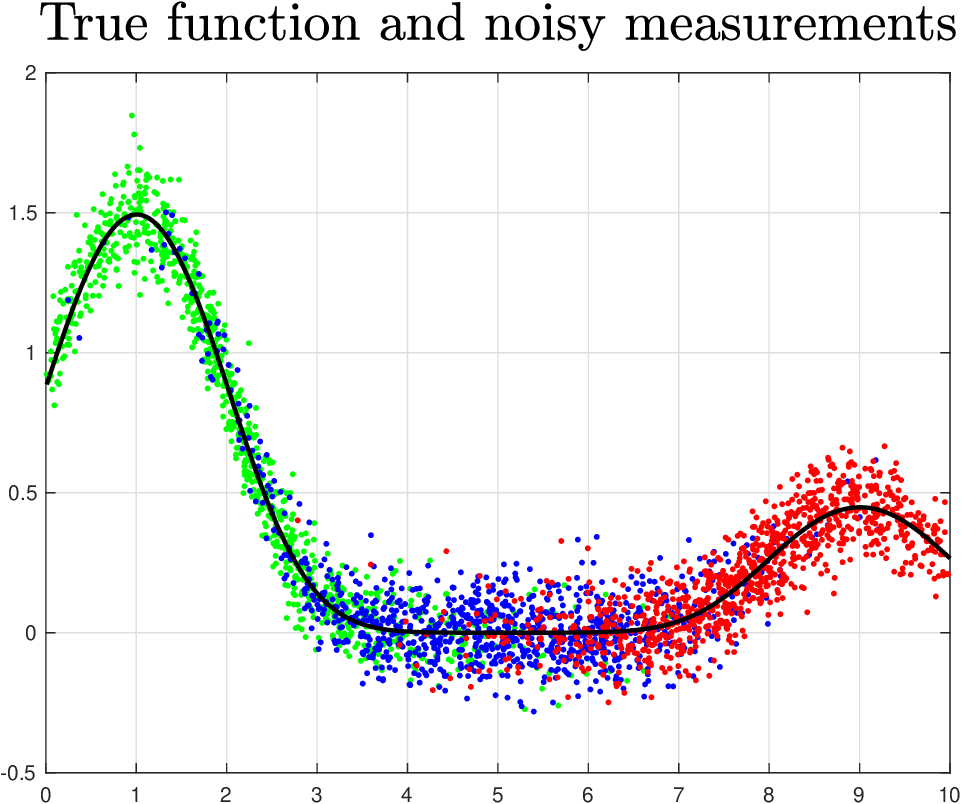}} \\
\caption{{\bf Left:} Sampling distribution adopted to generate the first set of data (green curve), the second (blue)
and the third (red). {\bf Right:} Regression function $\mu$ (black) and the three sets of 1000 output data generated using the sampling distributions displayed in the left panel.}
\label{Fig1}
\end{figure*}

\begin{figure*}
\center {\includegraphics[scale=0.42]{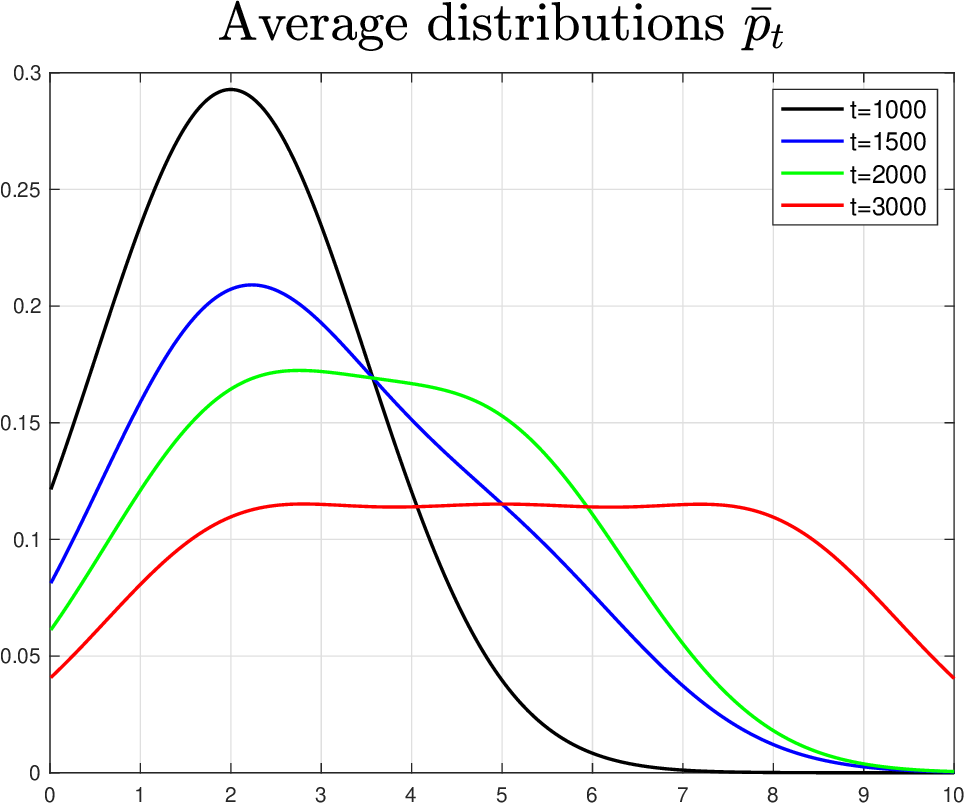}} \  {\includegraphics[scale=0.4]{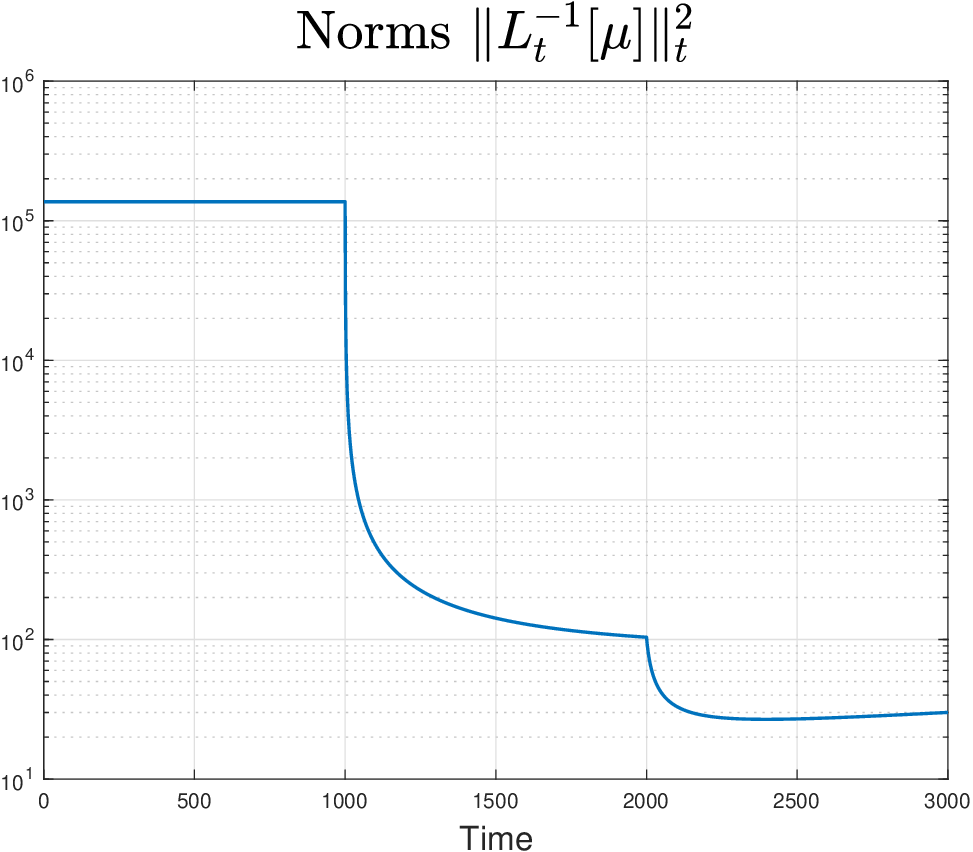}} \\
\caption{{\bf Left:} Average probability density $\bar{p}_t$ at four different time instants $t$. {\bf Right:} 
Norms $\|L_t^{-1}[\mu]\|^2_t$ as a function of $t$. Smaller values improve the learning rate.}
\label{Fig2}
\end{figure*}

\begin{figure*}
\center {\includegraphics[scale=0.55]{
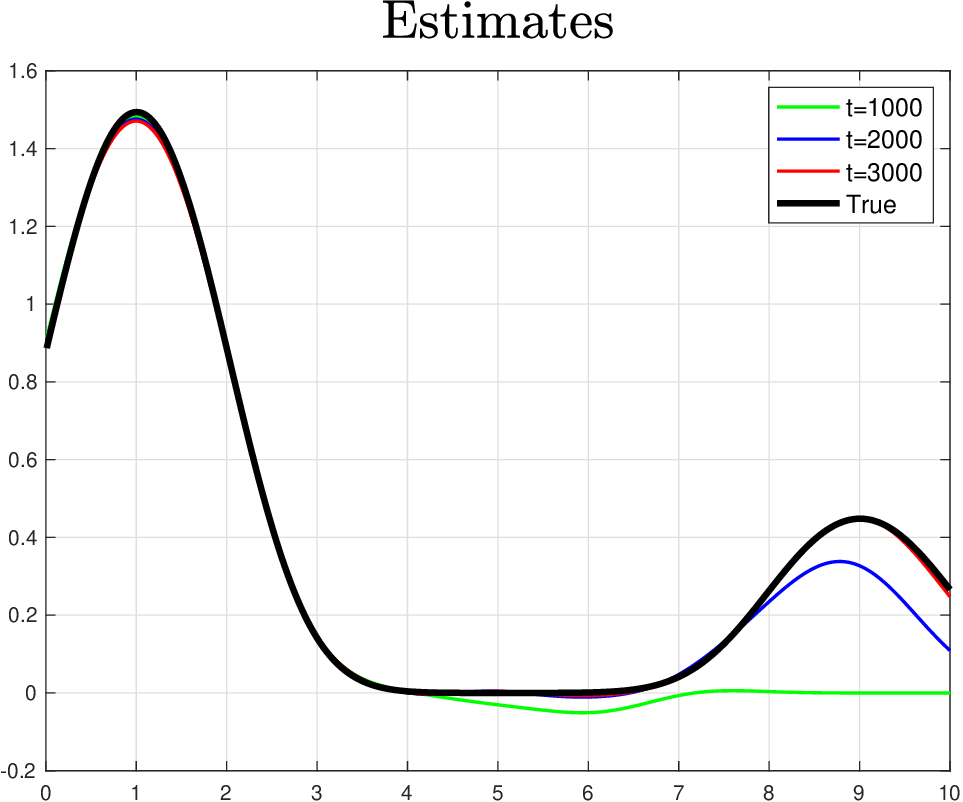}} \\
\caption{Regression function $\mu$ (black) and estimates obtained at time instants $t=1000,2000,3000$.}
\label{Fig3}
\end{figure*}

We consider an experiment where an unknown function $\mu$ 
has to be reconstructed by an agent over the interval
$[0,10]$ from 3000 input-output data. 
Specifically, let $K$ be the Gaussian kernel given by
\begin{equation}\label{TheGK}
K(x,a)=\exp(-(x-a)^2)
\end{equation}
and define
\begin{eqnarray}
    h(x) &=&  \begin{cases}
   1& 0 \leq x \leq 2, \\
    0.3 & x \geq 8, \\
    0 & \text{elsewhere.}
    \end{cases}\label{Functionh}
\end{eqnarray}
Then, the regression function is
\begin{equation}\label{TheRG}
\mu(x)= \int_0^{10} K(x,a)h(a)da
\end{equation}
and it is displayed in the right panel of Fig. \ref{Fig1} (black line). 
The input locations $x_i$ are independently generated adopting  
different probability densities. In particular, the set $\mathcal{P}$ contains the 
three truncated Gaussians displayed in the left panel of Fig. \ref{Fig1}. 
They regulate the agent's movements while the collected output data $y_i$ follow
\eqref{MM} with the $\nu_i$ forming a white Gaussian noise of variance $\sigma^2=0.01$.
The first density, represented by the green curve in left panel of Fig. \ref{Fig1}, 
is used to generate the first 1000 input locations, i.e. the set $\{x_i\}_{i=1}^{1000}$. The 
corresponding output data are the green points in the right panel of the same figure. 
One can see that this first distribution allows the agent to explore mainly the left side of the domain.
One can then think that
external directives drive the agent towards the right part of the region. Hence, its movements 
are regulated by the other two probability density functions
plotted in the left panel of Fig. \ref{Fig1}: the input locations $\{x_i\}_{i=1001}^{2000}$ 
are drawn from the blue curve while the last set $\{x_i\}_{i=2001}^{3000}$ is generated using the red one.
In this way, other 2000 output samples are collected, visible as blue and red points in the right panel.\\
To apply Theorem \ref{Main}, we have first to check that the assumptions 
reported in Section \ref{Assum} hold true.
Assumption \ref{A1} is trivially satisfied since the input locations, even if generated by a nonstationary stochastic process,
are all mutually independent. For what regards Assumption \ref{A2}, from 
the structure of the regression function $\mu$, it is easy to obtain 
\begin{equation}\label{hStruct}
L_p^{-1}[\mu](x)=\frac{h(x)}{p(x)} \qquad \forall p \in co\mathcal{P}.
\end{equation}
In view of the nature of the set $\mathcal{P}$, it is also immediate to check that   
$$
0<A \leq p(x) \leq B<\infty  \qquad \forall x \in [0,10], p \in co\mathcal{P}.
$$
This ensures the fullfilment of Assumption \ref{A2}, in particular of
\eqref{A2Cond} with $r=-1$.
The evolution of the average distributions $\bar{p}_t$ is displayed in the left panel of Fig. \ref{Fig2}. 
Note that as the time $t$ increases, such distribution is able to cover better (more uniformly) the function domain.
Through \eqref{hStruct}, it is also easy to 
compute 
the norms $\| L_t^{-1} [\mu]  \|_{t}$ which depend on $\bar{p}_t$.
In fact, one has
 $$
 \| L_t^{-1} [\mu]  \|^2_{t} = \int_0^{10} \frac{h^2(x)}{\bar{p}_t(x)} dx
 $$
 and the integral can be calculated numerically with high precision.
Their evolution is displayed in the right panel of the same figure where one can see
how their values quickly decrease after the instant $t=1000$, i.e. when the first sampling distribution
is replaced by the second one.
In view of \eqref{FirstTermB}, this is related to improvement of the learning rate. This 
phenomenon is clearly visible
in Fig. \ref{Fig3} which plots function estimates
$\hat{\mu}_t$, for $t=1000,2000$ and $3000$, obtained using \eqref{Tikhonov}
with the Gaussian kernel \eqref{TheGK} and 
$$\gamma=\frac{\sigma^2}{t^{0.25}}.$$
Using the first sampling distribution, i.e. up to instant $t=1000$, the recontruction of the
right part of $\mu$ is poor while the sup-norm of the estimation error goes quickly to zero for
$t>1000$, i.e. when
the other sampling distributions are adopted.

\section{Conclusions}

In many real applications, 
agents are distributed in an environment with the aim of reconstructing
a sensorial field. Ofen, it can be convenient to update on-line their movements rules 
e.g. to exploit the acquired knowledge on the 
unknown function or to follow external directives that have detected 
significant events in specific regions.
These tasks, which include also the important exploration-exploitation
problems, go beyond standard  machine learning since they require to apply 
supervised learning in dynamic and non stationary settings.
This problem has been here faced through the analysis of kernel ridge regression 
with input locations drawn from 
sampling distributions that may freely change in time.
Our main result thus provides conditions which ensure
convergence to the optimal predictor and learning rates in rather general settings, 
including situations where stochastic adaptation
may occur infinitely often.

\end{document}